\documentclass{article}
\usepackage{apacite}

\usepackage[preprint]{nips_2018}

\usepackage[utf8]{inputenc} 
\usepackage[T1]{fontenc}    
\usepackage{hyperref}       
\usepackage{url}            
\usepackage{booktabs}       
\usepackage{amsfonts}       
\usepackage{nicefrac}       
\usepackage{microtype}      
\usepackage{amsmath} 
\usepackage{amssymb}
\usepackage{graphicx}
\usepackage{subfigure}
\usepackage{geometry}

\newtheorem{lemma}{Lemma}

\newtheorem{definition}{Definition}
\newtheorem{conj}{Conjecture}

\usepackage{mathrsfs}

\title{GAN-QP: A Novel GAN Framework without Gradient Vanishing and Lipschitz Constraint}

\author{
  Jianlin Su \\
  School of Mathematics\\
  Sun Yat-sen University\\
  Guangdong, China \\
  \texttt{bojone@spaces.ac.cn} \\
}

\begin{document}

\maketitle

\begin{abstract}
  We know SGAN may have a risk of gradient vanishing. A significant improvement is WGAN, with the help of 1-Lipschitz constraint on discriminator to prevent from gradient vanishing. Is there any GAN having no gradient vanishing and no 1-Lipschitz constraint on discriminator? We do find one, called GAN-QP.

  To construct a new framework of Generative Adversarial Network (GAN) usually includes three steps: 1. choose a probability divergence; 2. convert it into a dual form; 3. play a min-max game. In this articles, we demonstrate that the first step is not necessary. We can analyse the property of divergence and even construct new divergence in dual space directly. As a reward, we obtain a simpler alternative of WGAN: GAN-QP. We demonstrate that GAN-QP have a better performance than WGAN in theory and practice.
\end{abstract}

\section{From Divergence to GAN}

\subsection{Divergence}

Most Generative Adversarial Networks (GANs, \cite{Goodfellow2014Generative}) are based on a certain form of probability divergence. A divergence is a function $\mathcal{D}$ of two variables $p,q$ satisfies the following definition:
\begin{definition}\label{def:div}
If $\mathcal{D}$ is function of two variables $p,q$ satisfies the following properties:
\begin{enumerate}
\item $\mathcal{D}[p, q]\geq 0$;
\item $p = q \Leftrightarrow \mathcal{D}[p, q]=0$.
\end{enumerate}
We say $\mathcal{D}$ is a divergence between $p$ and $q$.
\end{definition}
Compared with the axiomatic defination of distance, a divergence do not need symmetry and triangle inequality necessarily. Divergence only keeps the fundamental property for measuring the difference between $p$ and $q$.

\subsection{Dual Form}

If $p,q$ represent two probability distributions, $\mathcal{D}[p, q]$ becomes a functional and we call it probability divergence. For example, we have Jensen-Shannon divergence (JS divergence):
\begin{equation}\mathcal{JS}[p(x), q(x)] = \frac{1}{2}\int p(x)\log\frac{p(x)}{\frac{1}{2}[p(x)+q(x)]}dx + \frac{1}{2}\int q(x)\log\frac{q(x)}{\frac{1}{2}[p(x)+q(x)]}dx\label{eq:js-def}\end{equation}

In most cases, we can find a dual form for a probability divergence \citep{Nowozin2016f}. For example, the dual form of JS divergence is
\begin{equation}\begin{aligned}\mathcal{JS}[p(x), q(x)] =& \max_T\, \frac{1}{2}\int p(x)\log \sigma(T(x)) dx + \frac{1}{2}\int q(x) \log (1 - \sigma(T(x))) dx + \log 2\\
=& \max_T\, \frac{1}{2}\mathbb{E}_{x\sim p(x)}[\log \sigma(T(x))] + \frac{1}{2}\mathbb{E}_{x\sim q(x)}[\log (1 - \sigma(T(x)))] + \log 2
\end{aligned}\label{eq:js-dual}\end{equation}
Here $\sigma(x)=1/(1+e^{-x})$ is sigmoid function. The dual form can convert the integral of original divergence into a sampling form, which allows us to estimate it by Monte Carlo method. That is the essential to GAN. The dual form always has a max operation, which means a divergence is a supremum of a family of functions.

\subsection{Min-Max Game}

With the help of dual form of probability divergence, we can train a generator to generate the distribution we are interested via playing a min-max game. For example, using $\eqref{eq:js-dual}$ we have
\begin{equation}G,T = \mathop{\arg\min}_G\mathop{\arg\max}_T\, \mathbb{E}_{x\sim p(x)}[\log \sigma(T(x))]+ \mathbb{E}_{x=G(z),z\sim q(z)}[\log (1 - \sigma(T(x)))]\label{eq:js-min-max}\end{equation}
For a fixed $T$, the goal of $G$ is
\begin{equation}\begin{aligned}G =& \mathop{\arg\min}_G \mathbb{E}_{x\sim p(x)}[\log \sigma(T(x))]+ \mathbb{E}_{x=G(z),z\sim q(z)}[\log (1 - \sigma(T(x)))]\\
=& \mathop{\arg\min}_G \mathbb{E}_{x=G(z),z\sim q(z)}[\log (1 - \sigma(T(x)))]\end{aligned}\end{equation}
However, the loss $\log (1 - \sigma(T(x)))$ is not always good for optimization, so we usually use a equivalent loss, such as $-\log \sigma(T(x))$ and $- T(x)$. Namely, we may adjust the loss of generator for a better optimization, rather than playing the original min-max game.

\section{Divergence in Dual Space}

\subsection{Steps to GAN}

From the above discussion, we can see that to construct a GAN usually includes three steps:
\begin{enumerate}
\item choose a probability divergence;
\item onvert it into a dual form;
\item play a min-max game.
\end{enumerate}
But we know that only the last two steps are useful for practice. The fisrt step is only a theoretical concept and is not very important for a GAN. Therefore, a natural thought is: why not analyse the property of divergence and even construct new divergence in dual space directly? Our following content will demonstrate this thought is a very simple approach to build and understand GANs.

\subsection{SGAN}

We start from the Standard GAN (SGAN, \cite{Goodfellow2014Generative}) as an example to show how we can achieve the goal. From the appendix $\ref{add:js}$, we have Lemma $\ref{lemma:js}$:
\begin{lemma}\label{lemma:js} The following $\mathcal{D}[p(x),q(x)]$ defines a probability divergence
\begin{equation}\mathcal{D}[p(x),q(x)] = \max_T\, \frac{1}{2}\mathbb{E}_{x\sim p(x)}[\log \sigma(T(x))] + \frac{1}{2}\mathbb{E}_{x\sim q(x)}[\log (1 - \sigma(T(x)))] + \log 2\label{eq:js-dual-2}\end{equation}
\end{lemma}
It is worth to highlight that we prove $\mathcal{D}[p(x),q(x)]$ is a probability divergence\footnote{Namely, satisfying the definition $\ref{def:div}$.} in dual space, not need the original defination $\eqref{eq:js-def}$. Getting rid of the original defination of divergence allows us to seek more powerful divergence in dual space.

Now we have a divergence $\mathcal{D}[p(x),q(x)]$ defined by Lemma $\ref{lemma:js}$, so we can train a generator by minimizing $\mathcal{D}[p(x),q(x)]$, which results in the min-max game $\eqref{eq:js-min-max}$.

The difficulty arises while there is almost no intersection between $p(x)$ and $q(x)$. For example, we consider $p(x)=\delta(x-\alpha),q(x)=\delta(x-\beta)$ and $\alpha\neq \beta$. Now we have
\begin{equation}\mathcal{D}[p(x),q(x)]=\max_T\,\frac{1}{2}\log\sigma(T(\alpha)) + \frac{1}{2}\log(1-\sigma(T(\beta)))+\log 2\end{equation}
because no constraint on $T$, we can let $T(\alpha)\to +\infty, T(\beta)\to -\infty$, to obain the maximum value of the above formula, that is
\begin{equation}\mathcal{D}[p(x),q(x)]=\log 2\end{equation}
So if there is almost no intersection between $p(x)$ and $q(x)$, this divergence of them is a constant $\log 2$, whose gradients are zeros. In this situation, generator can not imporve via gradient descent method. And we know this situation will happen with very high probability \citep{Arjovsky2017Towards}. Therefore it's hard to train a good generative model under the framework of SGAN.

These conclusions can be popularized to any kind of $f$-GANs \citep{Nowozin2016f} in parallel, including LSGAN \citep{Mao2016Least}. And all of them suffer the same difficulty.

\subsection{WGAN}

We turn to a new kind of divergence by Lemma $\ref{lemma:wd}$:
\begin{lemma}\label{lemma:wd} The following $\mathcal{W}[p(x),q(x)]$ defines a probability divergence
\begin{equation}\mathcal{W}[p(x),q(x)] = \max_{T,\,\Vert T\Vert_L \leq 1}\, \mathbb{E}_{x\sim p(x)}[T(x)] - \mathbb{E}_{x\sim q(x)}[T(x)]\label{eq:wd-dual}\end{equation}
here
\begin{equation}\Vert T\Vert_L = \max_{x\neq y} \frac{|T(x)-T(y)|}{d(x,y)}\end{equation}
and $d(x, y)$ is any distance metric of $x,y$. $d(x,y)=\Vert x-y\Vert_1, d(x,y)=\Vert x-y\Vert_2$ is frequently-used.
\end{lemma}
The proof is in appendix $\ref{add:wd}$. Interestingly, the proof is very simple compared the corresponding proof of $f$-divergence.

Now we can play a new min-max game:
\begin{equation}G,T = \mathop{\arg\min}_G\mathop{\arg\max}_{T,\,\Vert T\Vert_L \leq 1}\, \mathbb{E}_{x\sim p(x)}[T(x)] - \mathbb{E}_{x=G(z),z\sim q(z)}[T(x)]\label{eq:wd-min-max}\end{equation}
That is what we call WGAN \citep{Arjovsky2017Wasserstein}.

Compared with $\eqref{eq:js-dual-2}$, $\mathcal{W}[p(x),q(x)]$ can reasonably measure the difference of $p(x),q(x)$ while they almost have no intersection. Let us consider $p(x)=\delta(x-\alpha),q(x)=\delta(x-\beta)$ and $\alpha\neq \beta$:
\begin{equation}\mathcal{W}[p(x),q(x)] = \max_{T,\,\Vert T\Vert_L \leq 1} T(\alpha) - T(\beta)\end{equation}
The constraint $\Vert T\Vert_L \leq 1$ means $|T(\alpha) - T(\beta)| \leq d(\alpha, \beta)$. So we have
\begin{equation}\mathcal{W}[p(x),q(x)] = d(\alpha,\beta)\end{equation}
The result is not a constant and its gradients are not zeros. So WGAN will not suffer gradient vanishing usually.

\subsection{WGAN-GP}

The essential problem of WGAN is how to constrain $T$ in $\Vert T\Vert_L \leq 1$, which currently has serveral solutions: weight clipping (WC, \cite{Arjovsky2017Wasserstein}), gradient penalty (GP, \cite{Gulrajani2017Improved}) and spectral normalization (SN, \cite{Miyato2018Spectral}).

Weight clipping is always unstable and has been abandoned in most cases. Spectral normalization is a better operation for not only WGANs but also many other GANs, but it constrains $T$ in a tiny subspace of $\Vert T\Vert_L \leq 1$, wasting the modeling power of $T$. 

It seems the best approch is gradient penalty now. Gradient penalty replaces $\Vert T\Vert_L$ with the norm of gradients $\Vert \nabla_x T\Vert$, and implement it as a penalty term:
\begin{equation}\begin{aligned}&T = \mathop{\arg\max}_T\, \mathbb{E}_{x\sim p(x)}[T(x)] - \mathbb{E}_{x\sim q(x)}[T(x)] - \lambda\mathbb{E}_{x\sim pq(x)}\left[(\Vert \nabla_x T(x)\Vert-1)^2\right]\\
&G = \mathop{\arg\min}_G\, \mathbb{E}_{x=G(z),z\sim q(z)}[-T(x)]
\end{aligned}\label{eq:wgan-gp}\end{equation}
whose $\lambda > 0$ is a hyperparameter and $pq(x)$ is the random linear interpolation of $p(x)$ and $q(x)$. It is called WGAN-GP.

WGAN-GP works well in many cases but it is just an empirical trick. Recently some researchs reveal some irrationality of WGAN-GP \citep{Wu2017Wasserstein}. Another disadvantage of WGAN-GP is the slow speed, while calculating the exact gradients needs more heavy computation.

\section{GAN with Quadratic Potential}

\subsection{A Quadratic Divergence}

From the discussion of SGAN and WGAN, we can see that an ideal divergence should not has any constraints on $T$ (like SGAN) and should give a reasonable measurement of $p(x),q(x)$ while they almost have no intersection (like WGAN). 

Here we propose such a divergence: 
\begin{lemma}\label{lemma:qp} The following $\mathcal{L}[p(x),q(x)]$ defines a probability divergence
\begin{equation}\begin{aligned}&\mathcal{L}[p(x),q(x)] \\
=& \max_{T}\, \mathbb{E}_{(x_r,x_f)\sim p(x_r)q(x_f)}\left[T(x_r,x_f)-T(x_f,x_r) - \frac{(T(x_r,x_f)-T(x_f,x_r))^2}{2\lambda d(x_r,x_f)}\right]\end{aligned}\label{eq:qp-dual}\end{equation}
$\lambda > 0$ is a hyperparameter and $d(x_r,x_f)$ is any distance metric of $x_r,x_f$. $d(x_r,x_f)=\Vert x_r-x_f\Vert_1, d(x_r,x_f)=\Vert x_r-x_f\Vert_2$ is frequently-used.
\end{lemma}
The proof is shown in appendix $\ref{add:qp}$. It is like a conventional divergence pluses a quadratic potential, so we call it quadratic potential divergence (QP-div).

Now we check $\mathcal{L}[p(x),q(x)]$ with $p(x)=\delta(x-\alpha),q(x)=\delta(x-\beta)$ to demenstrate QP-div will have a reasonable performance to two extreme distribution:
\begin{equation}\mathcal{L}[p(x),q(x)] = \max_{T}\, T(\alpha,\beta)-T(\beta,\alpha) - \frac{(T(\alpha,\beta)-T(\beta,\alpha))^2}{2\lambda d(\alpha,\beta)}\end{equation}
Let $z = T(\alpha,\beta)-T(\beta,\alpha)$, that is only a maximum value problem of quadratic function $z - \frac{z^2}{2\lambda d(\alpha,\beta)}$. We know it is $\frac{1}{2}\lambda d(\alpha,\beta)$, so
\begin{equation}\mathcal{L}[p(x),q(x)] = \frac{1}{2}\lambda d(\alpha,\beta)\end{equation}
We can see that $\mathcal{L}[p(x),q(x)]$ has similar property like $\mathcal{W}[p(x),q(x)]$, but with no constraints on $T$. It is very friendly to pratice.

\subsection{From QP-div to GAN-QP}

In theory, we can play a min-max game on QP-div to train a generative model
\begin{equation}\begin{aligned}&G,T=\mathop{\arg\min}_G \mathop{\arg\max}_T\, \mathbb{E}_{(x_r,x_f)\sim p(x_r)q(x_f)}\left[\mathscr{L}(x_r, x_f)\right] \\
&\mathscr{L}(x_r, x_f) = T(x_r,x_f)-T(x_f,x_r) - \frac{(T(x_r,x_f)-T(x_f,x_r))^2}{2\lambda d(x_r,x_f)}
\end{aligned}\end{equation}
However, $\mathscr{L}(x_r, x_f)$ is not a good loss for generator because there is a $d(x_r,x_f)$ in the denominator. Generator wants to minmize $\mathscr{L}(x_r, x_f)$, which will minimize $d(x_r,x_f)$ correspondingly. And we know any ready-to-use distance may not be used as a perfect metric of two samples.

We find that using $T(x_r,x_f)-T(x_f,x_r)$ as the loss of generator is enough. That results the following generative model called GAN with Quadratic Potential (GAN-QP):
\begin{equation}\begin{aligned}&T= \mathop{\arg\max}_T\, \mathbb{E}_{(x_r,x_f)\sim p(x_r)q(x_f)}\left[\mathscr{L}(x_r, x_f)\right] \\
&G = \mathop{\arg\min}_G\,\mathbb{E}_{(x_r,x_f)\sim p(x_r)q(x_f)}\left[T(x_r,x_f)-T(x_f,x_r)\right]
\end{aligned}\label{eq:gan-gp-gd}\end{equation}
Futher discussion can be found in appendix $\ref{add:ganqp}$.

\subsection{BiGAN-QP}

Like BiGAN \citep{Donahue2016Adversarial} or ALI \citep{Dumoulin2016Adversarially}, we can add an inference model $E(x)$ into GAN-QP. Only need to replace $x_r$ with $(x_r, E(x_r))$ and replace $x_f$ with $(G(z), z)$, we have
\begin{equation}\begin{aligned}T&= \mathop{\arg\max}_T\, \mathbb{E}_{x\sim p(x), z\sim q(z)}\left[\Delta T - \frac{\Delta T^2}{2\lambda d\big(x,E(x);G(z),z\big)}\right] \\
G &= \mathop{\arg\min}_{G,E}\,\mathbb{E}_{x\sim p(x), z\sim q(z)}\left[\Delta T\right]\\
\Delta T &= T\big(x,E(x);G(z),z\big)-T\big(G(z),z;x,E(x)\big)
\end{aligned}\label{eq:bigan-gp-gd}\end{equation}
That is what we called BiGAN-QP. In theory, $\eqref{eq:bigan-gp-gd}$ is enough for training the whole model. But actually it is hard to converge. We need to add some \emph{guiding term} to guide the training process. We can use reconstruction mse loss as the guiding term:
\begin{equation}\begin{aligned}T&= \mathop{\arg\max}_T\, \mathbb{E}_{x\sim p(x), z\sim q(z)}\left[\Delta T - \frac{\Delta T^2}{2\lambda d\big(x,E(x);G(z),z\big)}\right] \\
G &= \mathop{\arg\min}_{G,E}\,\mathbb{E}_{x\sim p(x), z\sim q(z)}\Big[\Delta T + \beta_1 \Vert z - E(G(z))\Vert^2 + \beta_2 \Vert x - G(E(x))\Vert^2\Big]\\
\Delta T &= T\big(x,E(x);G(z),z\big)-T\big(G(z),z;x,E(x)\big)
\end{aligned}\label{eq:bigan-gp-gd-2}\end{equation}
In practice, we find the following altering can improve quality of reconstruction:
\begin{equation}\begin{aligned}T&= \mathop{\arg\max}_T\, \mathbb{E}_{x\sim p(x), z\sim q(z)}\left[\Delta T - \frac{\Delta T^2}{2\lambda d\big(x,E(x);G(z),z\big)}\right] \\
G &= \mathop{\arg\min}_{G,E}\,\mathbb{E}_{x\sim p(x), z\sim q(z)}\Big[\Delta T + \beta_1 \Vert z - E(G_{ng}(z))\Vert^2 + \beta_2 \Vert x - G(E_{ng}(x))\Vert^2\Big]\\
\end{aligned}\label{eq:bigan-gp-gd-2}\end{equation}
where $G_{ng}$ and $E_{ng}$ means we stop its gradient at its output.

\section{Experiments}

\subsection{Experimential Details}

Our experiments are mainly conducted on CelebA HQ dataset \citep{Karras2017Progressive}. Our basic setup follows DCGANs \citep{Radford2015Unsupervised}, and is implemented in Keras \citep{chollet2015keras}, and available in my repository\footnote{\url{https://github.com/bojone/gan-qp}}. We use the Adam optimizer \citep{Kingma2014Adam}, with a constant learning rate of $2\times 10^{-4}$ and $\beta_1 = 0.5, \beta_2 = 0.999$ in both $T$ and $G$. We train GAN-QP with two D steps per G step.

In $\eqref{eq:gan-gp-gd}$, discriminator $T$ is a model with both real sample $x_r$ and fake sample $x_f$ as inputs. But in our experiments, we find just using one sample as input has generated good performance. In other words, $T(x_r,x_f)\equiv T(x_r)$ is enough. We try the architecture like $T(x_r, x_f)=D([E(x_r), E(x_f)])$ but there is no obvious improvement. So the final loss we use is
\begin{equation}\begin{aligned}&T= \mathop{\arg\max}_T\, \mathbb{E}_{(x_r,x_f)\sim p(x_r)q(x_f)}\left[T(x_r)-T(x_f) - \frac{(T(x_r)-T(x_f))^2}{2\lambda d(x_r,x_f)}\right] \\
&G = \mathop{\arg\min}_G\,\mathbb{E}_{(x_r,x_f)\sim p(x_r)q(x_f)}\left[T(x_r)-T(x_f)\right]
\end{aligned}\end{equation}

Hyperparameter $\lambda$ is
\begin{equation}\lambda = \left\{\begin{aligned}&\frac{10}{w\times h\times ch},\,\text{while } d(x_r,x_f)=\Vert x_r-x_f\Vert_1\\
&\frac{10}{\sqrt{w\times h\times ch}},\,\text{while } d(x_r,x_f)=\Vert x_r-x_f\Vert_2  \end{aligned}
\right.\end{equation}
$w,h,ch$ is the width, height and the number of channels of the input images.We test both L1 norm $d(x_r,x_f)=\Vert x_r-x_f\Vert_1$ and L2 norm $d(x_r,x_f)=\Vert x_r-x_f\Vert_2$ in our experiments, but they have no significant statistical difference.

The quantitative index we use to evaluate a GAN is Frechet Inception Distance (FID, \cite{Heusel2017GANs}). We also re-implement it in Keras.

\subsection{Basic Comparison}

\begin{figure}[h]
  \centering
  \includegraphics[width=10cm]{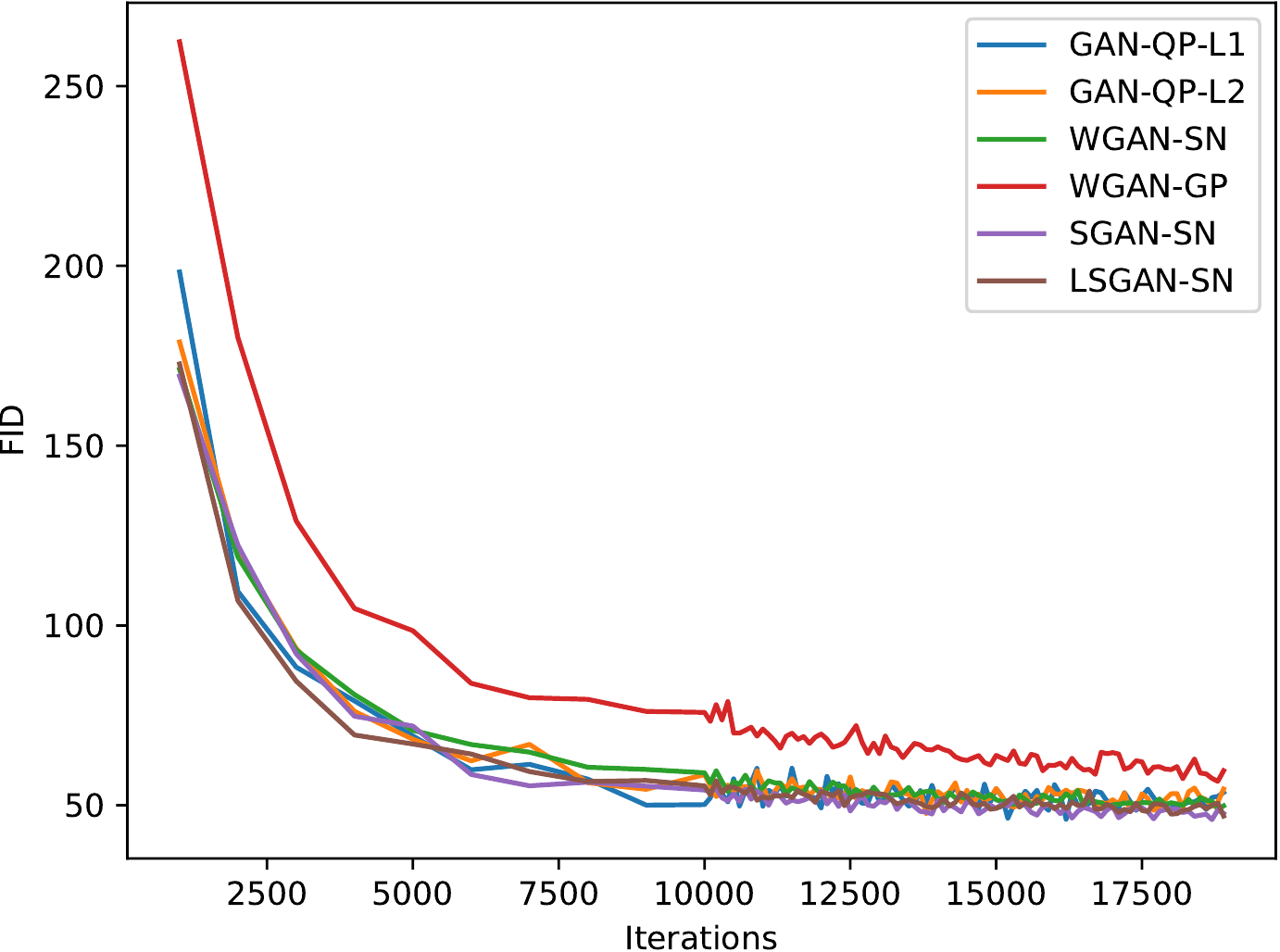}
  \caption{FID comparison of GAN-QP with L1/L2 distance and WGAN-GP,WGAN-SN,SGAN-SN,LSGAN-SN. WGAN-GP is generally worse than others. The best two is GAN-QP and SGAN-SN.}
  \label{fig:fid-comparison}
\end{figure}

Firstly, we compared GAN-QP with WGAN-GP, WGAN-SN (WGAN with spectral normalization), SGAN-SN (SGAN with spectral normalization), LSGAN-SN (LSGAN with spectral normalization) on 128x128 resolution. $\lambda$ in WGAN-GP $\eqref{eq:wgan-gp}$ we use is 10. Then Batch Normalization is removed from discriminator of WGAN-GP and other hyperparameters are as the same as GAN-QP. Each experiment is repeated twice for obtaining more reliable conclusion. The comparison is shown in Figure $\ref{fig:fid-comparison}$ and Table $\ref{tab:128-3}$. 

We can see that there is no obvious difference between GAN-QP-L1 and GAN-QP-L2, which means GAN-QP is robust to distance metric. The best two results come from GAN-QP and WGAN-SN. The worst is WGAN-GP. Generally, the FID curve of WGAN-SN and SGAN-SN is more smoother and GAN-QP is more shaking. But GAN-QP keeps the best performance as same as SGAN-SN.

\begin{table}[h]
  \renewcommand\arraystretch{1.5}
  \caption{Final performance of GANs on 128x128 resolution.}
  \label{tab:128-3}
  \centering
  \begin{tabular}{c|ccccc}
    \hline
    \hline
    & GAN-QP-L1 / L2 & WGAN-GP & WGAN-SN & SGAN-SN & LSGAN-SN  \\
    \hline
   Best FID & 45.0 / 44.7 & 55.5 & 47.8 & 44.5 & 45.8\\
   \hline
   Speed & 1x / 1x & 1.5x & 1x & 1x & 1x\\
    \hline
    \hline
  \end{tabular}
\end{table}

\subsection{Higher Resolution}

\begin{figure}[h]
  \centering
  \includegraphics[width=12cm]{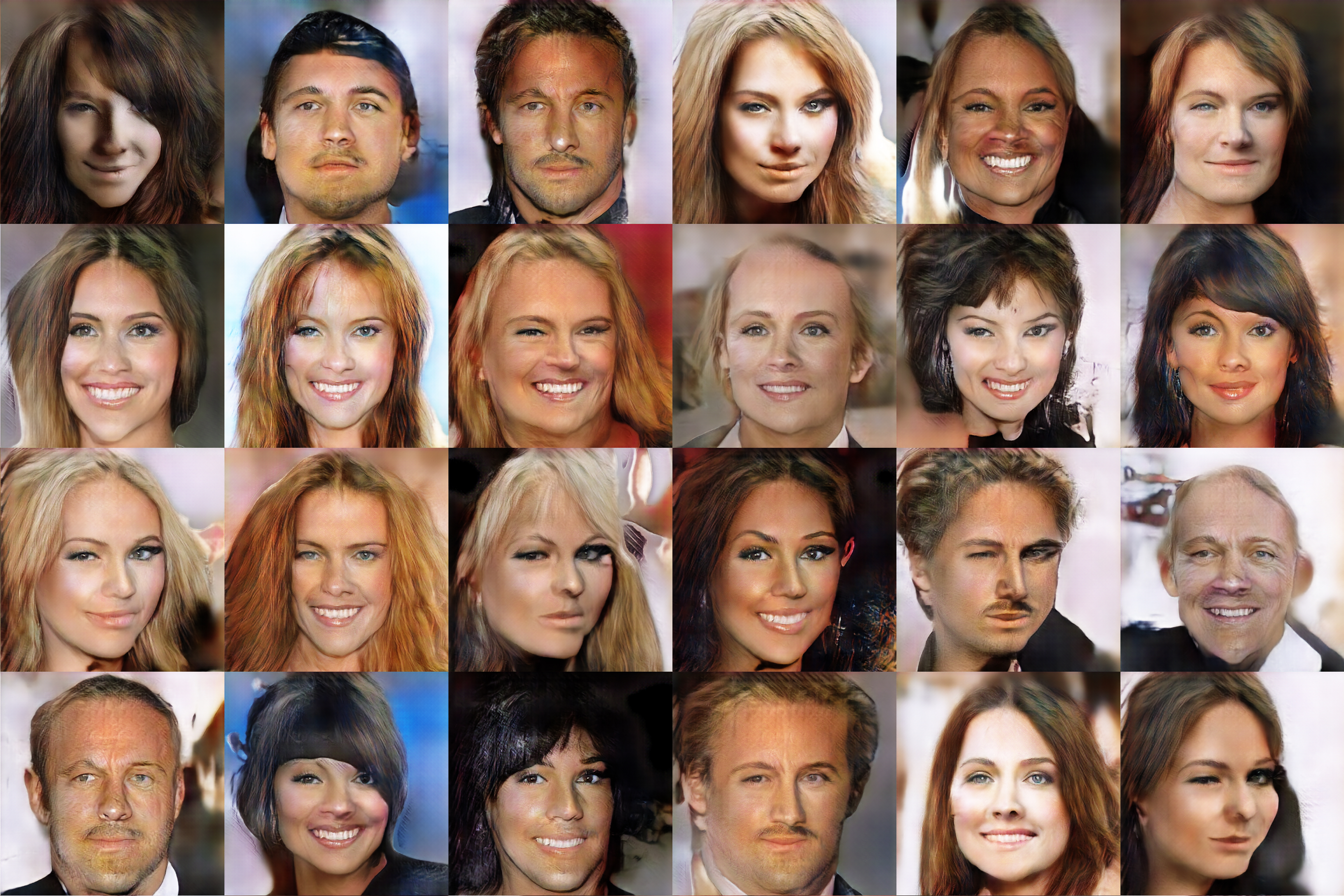}
  \caption{Random samples from GAN-QP on 512x512 resolution. The final FID is 26.64. And it costs 2 days to finish training on one gtx 1080ti.}
  \label{fig:512}
\end{figure}

On 128x128 resolution, SGAN-SN and GAN-QP has the same best performance. If we turn to 256x256 resolution, we can see that GAN-QP achieves a better FID than SGAN-SN (Table $\ref{tab:256-3}$). It even works well on 512x512 resolution (Figure $\ref{fig:512}$).

\begin{table}[h]
  \renewcommand\arraystretch{1.5}
  \caption{Final performance of GAN-QP and SGAN-SN on 256x256 resolution.}
  \label{tab:256-3}
  \centering
  \begin{tabular}{c|cc}
    \hline
    \hline
    & GAN-QP & SGAN-SN  \\
    \hline
   Best FID & 22.7 & 27.9\\
    \hline
    \hline
  \end{tabular}
\end{table}

\subsection{Reconstruction of BiGAN-QP}

\begin{figure}[h]
  \centering
  \includegraphics[width=13.6cm]{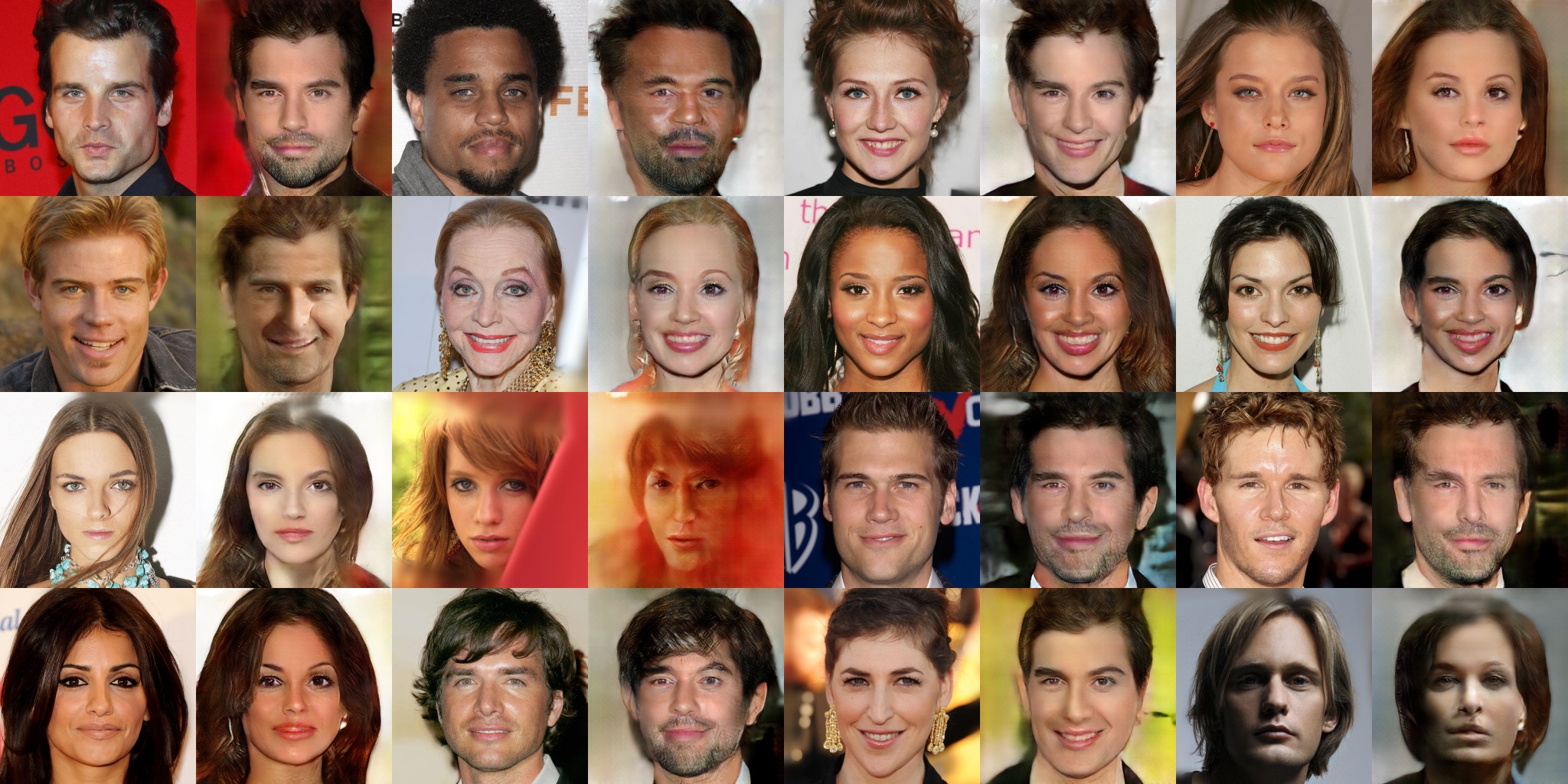}
  \caption{Reconstruction(right) of real images(left) on 256x256 resolution.}
  \label{fig:ae}
\end{figure}

We also use a simpler version of $\eqref{eq:bigan-gp-gd-2}$ to evaluate the performance of BiGAN-QP:
\begin{equation}\begin{aligned}T&= \mathop{\arg\max}_T\, \mathbb{E}_{x\sim p(x), z\sim q(z)}\left[\Delta T  - \frac{\Delta T^2}{2\lambda d\big(x,E(x);G(z),z\big)}\right] \\
G &= \mathop{\arg\min}_{G,E}\,\mathbb{E}_{x\sim p(x), z\sim q(z)}\Big[\Delta T + \beta_1 \Vert z - E(G_{ng}(z))\Vert^2 + \beta_2 \Vert x - G(E_{ng}(x))\Vert^2\Big]\\
\Delta T &= T(x,E(x))-T(G(z),z)
\end{aligned}\label{eq:bigan-gp-gd-3}\end{equation}
while $\beta_1 = 4 / \dim_z$ and $\beta_2 = 6 / \dim_x$. The result is shown in Figure $\ref{fig:ae}$.

\section{Conclusion}

In this paper, we demonstrate that we can explore probability divergence directly, which is more convenient and flexible. Starting from this idea, we find out a novel divergence called QP-div, which has excellent characteristics, does not require the 1-Lipschitz constraint and does not require the extract gradient penalty. As a concrete example, we construct a new framework of GAN equiped QP-div: GAN-QP. And the experiments demonstrate the stability and superiority of GAN-QP.

\newpage
\medskip


\bibliographystyle{apacite}
\bibliography{ganqp.bib}

\newpage
\appendix

\begin{center}
\textbf{\Large Supplemental Materials}
\end{center}

\section{Detailed Derivation}

\subsection{$\eqref{eq:js-dual-2}$ is a divergence\label{add:js}}

Firstly, it is trivial to see that $\mathcal{D}[p(x), q(x)]$ is nonnegative since we can always let $T(x)\equiv 0$:
\begin{equation}\begin{aligned}\mathcal{D}[p(x),q(x)] =& \max_T\, \frac{1}{2}\mathbb{E}_{x\sim p(x)}[\log \sigma(T(x))] + \frac{1}{2}\mathbb{E}_{x\sim q(x)}[\log (1 - \sigma(T(x)))] + \log 2\\
\geq& \frac{1}{2}\mathbb{E}_{x\sim p(x)}[\log \sigma(0)] + \frac{1}{2}\mathbb{E}_{x\sim q(x)}[\log (1 - \sigma(0))] + \log 2\\
=& 0
\end{aligned}\end{equation}

Next, we show $\mathcal{D}[p(x), p(x)]=0$, which is also simple:
\begin{equation}\begin{aligned}\mathcal{D}[p(x),p(x)] =& \max_T\, \frac{1}{2}\mathbb{E}_{x\sim p(x)}[\log \sigma(T(x)) + \log (1 - \sigma(T(x)))] + \log 2\\
=& \max_T\, \frac{1}{2}\mathbb{E}_{x\sim p(x)}[\log \big(\sigma(T(x))(1 - \sigma(T(x)))\big)] + \log 2
\end{aligned}\end{equation}
It is not difficult to prove that the maximum value of $\sigma(z)(1-\sigma(z))$ is $\frac{1}{4}$ at $z=0$, so we have
\begin{equation}\mathcal{D}[p(x),p(x)] = \frac{1}{2}\mathbb{E}_{x\sim p(x)}[\log \frac{1}{4}] + \log 2 = 0\end{equation}

Finally, we show $\mathcal{D}[p(x), q(x)] > 0$ if $p(x)\neq q(x)$.\footnote{Strictly, $p(x)\neq q(x)$ is not enough. The sufficient condition is $\int_{\{x|p(x)\neq q(x)\}}dx > 0$.} Let
\begin{equation}\sigma(T(x))=\frac{p(x)}{p(x)+q(x)}=\lambda(x)\end{equation}
we have
\begin{equation}\begin{aligned}\mathcal{D}[p(x),q(x)]=&\frac{1}{2}\int \left[p(x)\log \frac{p(x)}{p(x)+q(x)} + q(x) \log \frac{q(x)}{p(x)+q(x)}\right]dx+\log 2\\
=&\int \left(\frac{p(x)+q(x)}{2}\right)\Big[\lambda(x)\log \lambda(x) + \big(1-\lambda(x)\big)\log \big(1-\lambda(x)\big)\Big]dx+\log 2\end{aligned}\end{equation}
Becausce of $p(x)\neq q(x)$, $\lambda(x)\not\equiv 1/2$, and we know $\lambda \log \lambda + (1-\lambda)\log(1-\lambda) < -\log 2$ if $\lambda\neq \frac{1}{2}$. Therefore,
\begin{equation}\mathcal{D}[p(x),q(x)] > \int \left(\frac{p(x)+q(x)}{2}\right)(-\log 2)dx+\log 2 = 0\end{equation}

\subsection{$\eqref{eq:wd-dual}$ is a divergence\label{add:wd}}

Firstly, it is trivial to see that $\mathcal{W}[p(x), q(x)]$ is nonnegative since we can always let $T(x)\equiv 0$:
\begin{equation}\mathcal{W}[p(x),q(x)] \geq \mathbb{E}_{x\sim p(x)}[0] - \mathbb{E}_{x\sim q(x)}[0] = 0\end{equation}

Next, $\mathcal{W}[p(x), p(x)]=0$ is also trivial. So we only need to show $\mathcal{W}[p(x), q(x)] > 0$ if $p(x)\neq q(x)$. It it actually not difficult because we only need to let $T_0(x) = \text{sign}(p(x) - q(x))$:
\begin{equation}\mathbb{E}_{x\sim p(x)}[T_0(x)] - \mathbb{E}_{x\sim q(x)}[T_0(x)] =\int (p(x)-q(x))\cdot\text{sign}(p(x) - q(x))dx > 0\end{equation}
That means $\mathcal{W}[p(x), p(x)] > 0$.

\subsection{$\eqref{eq:qp-dual}$ is a divergence\label{add:qp}}

Firstly, it is trivial to see that $\mathcal{W}[p(x), q(x)]$ is nonnegative since we can always let $T(x_r, x_f)\equiv 0$:
\begin{equation}\mathcal{L}[p(x),q(x)] \geq \mathbb{E}_{(x_r,x_f)\sim p(x_r)q(x_f)}\left[0 - \frac{0^2}{2\lambda d(x,y)}\right]=0\end{equation}

Next, we have
\begin{equation}\begin{aligned}\mathcal{L}[p(x),p(x)] = & \max_{T}\, \mathbb{E}_{(x_r,x_f)\sim p(x_r)p(x_f)}\left[-\frac{(T(x_r,x_f)-T(x_f,x_r))^2}{2\lambda d(x,y)}\right]\end{aligned}\end{equation}
Obviously, the maximum value is zero. So $\mathcal{L}[p(x),p(x)]=0$.

Finally, if $p(x)\neq q(x)$, we let
\begin{equation}T_0(x_r, x_f) = \text{sign}(p(x_r)q(x_f) - p(x_f)q(x_r))\end{equation}
now we have
\begin{equation}\begin{aligned}\gamma_1 = &\mathbb{E}_{(x_r,x_f)\sim p(x_r)q(x_f)}\left[T_0(x_r,x_f)-T_0(x_f,x_r)\right] \\
=&\iint p(x_r)q(x_f) \left[T_0(x_r,x_f)-T_0(x_f,x_r)\right] dx_r dx_f\\
=&\iint \left[p(x_r)q(x_f) - p(x_f)q(x_r)\right] T_0(x_r,x_f) dx_r dx_f\\
=&\iint \left[p(x_r)q(x_f) - p(x_f)q(x_r)\right] \cdot \text{sign}(p(x_r)q(x_f) - p(x_f)q(x_r)) dx_r dx_f > 0\\
\gamma_2=&\mathbb{E}_{(x_r,x_f)\sim p(x_r)q(x_f)}\left[\frac{(T_0(x_r,x_f)-T_0(x_f,x_r))^2}{2\lambda d(x,y)}\right] \geq 0\end{aligned}\end{equation}
If $\gamma_1 > \gamma_2$, then
\begin{equation}\begin{aligned}&\mathbb{E}_{(x_r,x_f)\sim p(x_r)q(x_f)}\left[T(x_r,x_f)-T(x_f,x_r) - \frac{(T(x_r,x_f)-T(x_f,x_r))^2}{2\lambda d(x,y)}\right]\\
=&\gamma_1 - \gamma_2 > 0\end{aligned}\end{equation}
else if $\gamma_1 \leq \gamma_2$, we can define
\begin{equation}T(x_r, x_f) = \frac{\gamma_1}{2\gamma_2}\cdot T_0(x_r, x_f)\end{equation}
then
\begin{equation}\begin{aligned}&\mathbb{E}_{(x_r,x_f)\sim p(x_r)q(x_f)}\left[T(x_r,x_f)-T(x_f,x_r) - \frac{(T(x_r,x_f)-T(x_f,x_r))^2}{2\gamma d(x,y)}\right]\\
=&\left(\frac{\gamma_1}{2\gamma_2}\right)\gamma_1 - \left(\frac{\gamma_1}{2\gamma_2}\right)^2\gamma_2 = \frac{\gamma_1^2}{4\gamma_2} > 0\end{aligned}\end{equation}

Therefore $\mathcal{L}[p(x),q(x)] > 0$.

\section{Analyse of GAN-QP\label{add:ganqp}}

\subsection{Optimum Solution of $\eqref{eq:qp-dual}$}

\begin{lemma}\label{lemma:opt} the optimum solution of $\eqref{eq:qp-dual}$ satisfies
\begin{equation}\frac{p(x_r)q(x_f) - p(x_f)q(x_r)}{p(x_r)q(x_f) + p(x_f)q(x_r)} = \frac{T(x_r,x_f)-T(x_f,x_r)}{\lambda d(x_r, x_f)}\label{eq:opt-t}\end{equation}
\end{lemma}

The proof is the basic application of variational method:
\begin{equation}\begin{aligned}&\delta\iint p(x_r)q(x_f)\left[T(x_r,x_f)-T(x_f,x_r) - \frac{(T(x_r,x_f)-T(x_f,x_r))^2}{2\lambda d(x_r,x_f)}\right]dx_r dx_f\\
=&\iint p(x_r)q(x_f)\Bigg[\delta T(x_r,x_f)-\delta T(x_f,x_r)\\
&\qquad\qquad\qquad\qquad- \frac{(T(x_r,x_f)-T(x_f,x_r))(\delta T(x_r,x_f)-\delta T(x_f,x_r))}{\lambda d(x_r,x_f)}\Bigg]dx_r dx_f\\
=&\iint \Bigg[p(x_r)q(x_f) - p(x_f)q(x_r) \\
&\qquad\qquad\qquad\qquad- \Big(p(x_r)q(x_f) + p(x_f)q(x_r)\Big)\frac{T(x_r,x_f)-T(x_f,x_r)}{\lambda d(x_r,x_f)}\Bigg]\delta T(x_r,x_f)dx_r dx_f\\
\end{aligned}\end{equation}
The formula in square brackets must be identically equal to zero. Therefore
\begin{equation}\frac{p(x_r)q(x_f) - p(x_f)q(x_r)}{p(x_r)q(x_f) + p(x_f)q(x_r)} = \frac{T(x_r,x_f)-T(x_f,x_r)}{\lambda d(x_r, x_f)}\end{equation}

\subsection{Adaptive Lipschitz Constraint}

From $\eqref{eq:opt-t}$, it is not difficult to prove that the optimum $T(x_r, x_f)$ satisfies
\begin{equation}-1 \leq \frac{T(x_r,x_f)-T(x_f,x_r)}{\lambda d(x_r, x_f)}\leq 1\end{equation}
In other words, the optimum $T(x_r, x_f)$ satisfies Lipschitz constraint automatically. Therefore we can say $\eqref{eq:qp-dual}$ is a divergence with adative Lipschitz constraint.

\subsection{The Divergence of Generator}

We use $T(x_r,x_f)-T(x_f,x_r)$ rather than the whole $\mathscr{L}(x_r, x_f)$ as the loss of generator in $\eqref{eq:gan-gp-gd}$. And we have solved the optimum solution of $\eqref{eq:qp-dual}$ in Lemma $\ref{lemma:opt}$. Then we can see the ultimate goal of generator to minimize is
\begin{equation}\lambda \iint p(x_r)q(x_f)\frac{p(x_r)q(x_f) - p(x_f)q(x_r)}{p(x_r)q(x_f) + p(x_f)q(x_r)} d(x_r, x_f) dx_r dx_f \label{eq:gan-qp-g-loss}\end{equation}
Now we have Lemma $\ref{lemma:opt-g}$:
\begin{lemma}\label{lemma:opt-g}
\begin{equation}\label{eq:opt-g-loss}\tilde{\mathcal{L}}[p(x),q(x)]=\iint p(x_r)q(x_f)\frac{p(x_r)q(x_f) - p(x_f)q(x_r)}{p(x_r)q(x_f) + p(x_f)q(x_r)} d(x_r, x_f) dx_r dx_f\end{equation}
is also a probability divergence of $p(x),q(x)$.
\end{lemma}

Actually Lemma $\ref{lemma:opt-g}$ is a conclusion of Cauchy–Schwarz inequality. Firstly we let
\begin{equation}\mu(x_r, x_f) = \frac{d(x_r, x_f)}{p(x_r)q(x_f) + p(x_f)q(x_r)} > 0\end{equation}
Then by Cauchy–Schwarz inequality we have
\begin{equation}\begin{aligned}&\left(\iint \mu(x_r, x_f) p^2(x_r) q^2(x_f) dx_r dx_f\right)^2\\
= &\left(\iint \left(\sqrt{\mu(x_r, x_f)} p(x_r) q(x_f)\right)^2 dx_r dx_f\right)\left(\iint \left(\sqrt{\mu(x_f, x_r)} p(x_f) q(x_r)\right)^2 dx_f dx_r\right)\\
\geq & \left(\iint \mu(x_r, x_f) p(x_r) q(x_f) p(x_f) q(x_r) dx_r dx_f\right)^2
\end{aligned}\end{equation}
So
\begin{equation}\tilde{\mathcal{L}}[p(x),q(x)] = \iint \mu(x_r, x_f) p(x_r) q(x_f) \Big(p(x_r) q(x_f) - p(x_f) q(x_r)\Big) \mu(x_r, x_f) dx_r dx_f\geq 0\end{equation}
Two sides are equal if and only if $\sqrt{\mu(x_r, x_f)} p(x_r) q(x_f)\equiv \sqrt{\mu(x_f, x_r)} p(x_f) q(x_r)$, which means $p(x)\equiv q(x)$. Therefore $\eqref{eq:opt-g-loss}$ is a probability divergence, which means to lower $T(x_r,x_f)-T(x_f,x_r)$ is actually to lower the difference between $p(x)$ and $q(x)$. The divergence is weighted by $d(x_r, x_f)$, forcing generator to focus on the sample pairs of larger distance, which is in line with our intuition.

\subsection{Performance while No Intersection}

We have shown the QP-div in Lemma $\ref{lemma:qp}$ also works well while there is no intersection between $p(x)$ and $q(x)$. But now we use $T(x_r,x_f)-T(x_f,x_r)$ as the loss of generator, corresponding to the new divergence $\tilde{\mathcal{L}}[p(x),q(x)]$. Therefore we have to check the performance of $\tilde{\mathcal{L}}[p(x),q(x)]$ with $p(x)=\delta(x-\alpha),q(x)=\delta(x-\beta)$. That is very easy:
\begin{equation}\begin{aligned}&\tilde{\mathcal{L}}[\delta(x-\alpha),\delta(x-\beta)]\\
=&\iint \delta(x_r-\alpha)\delta(x_f-\beta)\frac{\delta(x_r-\alpha)\delta(x_f-\beta) - \delta(x_f-\alpha)\delta(x_r-\beta)}{\delta(x_r-\alpha)\delta(x_f-\beta) + \delta(x_f-\alpha)\delta(x_r-\beta)} d(x_r, x_f) dx_r dx_f\\
=&\frac{\delta(0)\delta(0) - \delta(\beta-\alpha)\delta(\alpha-\beta)}{\delta(0)\delta(0) + \delta(\beta-\alpha)\delta(\alpha-\beta)} d(\alpha, \beta)
\end{aligned}\end{equation}
We know $\delta(0)\to\infty$ and $\delta(\alpha-\beta)=0$ for $\alpha\neq\beta$, so the result is $d(\alpha, \beta)$, a reasonable measurement actually.

\subsection{Robustness of $\lambda$}

$\eqref{eq:gan-qp-g-loss}$ showed that $\lambda$ is just a scaler for $\tilde{\mathcal{L}}[p(x),q(x)]$. That means GAN-QP is insensitive to the hyperparameter $\lambda$, which is different from WGAN-GP. We only need to choose a suitable $\lambda$ to make the loss more readable (not very large and not very small).

\section{Future Work}

\subsection{A Conjecture}

Inspired by the form of QP-div $\eqref{eq:qp-dual}$, it may be extended as
\begin{conj}\label{conj:div}If
\begin{equation}\mathop{\arg\max}_{T}\,\mathbb{E}_{(x_r,x_f)\sim p(x_r)q(x_f)}\left[f(T(x_r,x_f))\right]\end{equation}
is a probability divergence of $p(x)$ and $q(x)$, so does
\begin{equation}\mathop{\arg\max}_{T}\,\mathbb{E}_{(x_r,x_f)\sim p(x_r)q(x_f)}\left[f(T(x_r,x_f)) - \frac{f^2(T(x_r,x_f))}{2\lambda d(x_r,x_f)}\right]\end{equation}
for some $\lambda$.\end{conj}

Conjecture $\ref{conj:div}$ means we can use
\begin{equation}\frac{f^2(T(x_r,x_f))}{2\lambda d(x_r,x_f)}\end{equation}
as a penalty term for any other GAN's discriminator to enhance the original GAN.

\subsection{Example}

For example, we can enhance SGAN $\eqref{eq:js-min-max},\eqref{eq:js-dual-2}$ with a quadratic potential term:
\begin{equation}\begin{aligned}&T =\mathop{\arg\max}_T\, \mathbb{E}_{(x_r,x_f)\sim p(x_r)q(x_f)}\left[f(x_r, x_f) - \frac{f^2(T(x_r,x_f))}{2\lambda d(x_r,x_f)}\right]\\
&G =\mathop{\arg\min}_G\, \mathbb{E}_{x_r\sim p(x_r), x_f = G(z), z\sim q(z)}\left[f(x_r, x_f)\right]\\
&f(x_r, x_f)=\frac{1}{2} \log \sigma(T(x_r)) + \frac{1}{2} \log (1 - \sigma(T(x_f))) + \log 2\end{aligned}\end{equation}
It has been validated preliminarily by our experiments. But it is still yield to be proven strictly.

\end{document}